\documentclass{article}
\usepackage{spconf,amsmath,epsfig}
\usepackage{diagbox}
\usepackage{pmat}
\usepackage{array}
\usepackage{booktabs}
\usepackage{multirow}
\usepackage{amssymb}

\title{PolSAR Image Classification based on
Polarimetric Scattering Coding and Sparse Support Matrix Machine}
\name{Xu Liu, Licheng Jiao, Dan Zhang, Fang Liu
\thanks{This work was supported in part by the State Key Program of National Natural Science of China (No.61836009, No. 91438201 and No. 91438103),
the National Natural Science Foundation of China (No. 61801351),
the National Science Basic Research Plan in Shaanxi Province of China (No.2018JQ6018),
the Fund for Foreign Scholars in University Research and Teaching Programs (the 111 Project) (No. B07048), the Fundamental Research Funds for the Central Universities (No. XJS17108) and the China Postdoctoral Fund (No. 2017M613081).}}

\address{Key Laboratory of Intelligent Perception and Image Understanding of Ministry of Education,\\
International Research Center for Intelligent Perception and Computation,\\
Joint International Research Laboratory of Intelligent Perception and Computation,\\
School of Artificial Intelligence, Xidian University, Xi¡¯an, Shaanxi Province 710071, China}
\begin{document}
%
\maketitle
\begin{abstract}
POLSAR image has an advantage over optical image because it can be acquired independently of cloud cover and solar illumination. PolSAR image classification is a hot and valuable topic for the interpretation of POLSAR image.
In this paper, a novel POLSAR image classification method is proposed based on polarimetric scattering coding and sparse support matrix machine. First, we transform the original POLSAR data to get a real value matrix by the polarimetric scattering coding, which is called polarimetric scattering matrix and is a sparse matrix.
Second, the sparse support matrix machine is used to classify the sparse polarimetric scattering matrix and get the classification map.
The combination of these two steps takes full account of the characteristics of POLSAR.
The experimental results show that the proposed method can get better results and is an effective classification method.
\end{abstract}
\section{Introduction}
\label{sec:intro}
Polarimetric synthetic aperture radar (PolSAR) images have been widely used in urban planning, agriculture assessment, environment monitoring and so on \cite{wang2017comparison}. These applications require the full understanding and interpretation of PolSAR images.

PolSAR image classification is an important and hot research topic. The classification is arranging the pixels to the different categories according to the certain rule. The common objects within the PolSAR images include land, buildings, water, sand, urban areas, vegetation, road, bridge and so on \cite{liuf2016hierarchical}.
In order to distinguish them, the features of the pixels should be fully extracted and mined. These images contain rich character of the target.
The feature extraction techniques can be divided into two kinds based on polarimetric characteristics: coherent target decomposition and incoherent target decomposition. The former acts on the scattering matrix to characterize completely polarized scattered waves, which contains the fully polarimetric information. The latter acts only on the mueller matrix, covariance matrix, or coherency matrix in order to characterize partially polarized waves \cite{dickinson2013classification}.

The coherent target decomposition algorithms mainly include the Pauli decomposition, the sphere-diplane-helix (SDH) decomposition, the symmetric scattering characterization method, Cameron decomposition, Yamaguchi Four-component scattering decomposition, General polarimetric model-based decomposition, and some advances.
The incoherent target decomposition algorithms mainly include Huynen decomposition, Freeman-Durden decomposition, Yamguchi four-component decomposition, Cloude-Pottier decomposition, etc \cite{Aghababaee2016Incoherent}.
In addition to feature based on the polarization mechanism,
there are some traditional features of natural images, which have been utilized to analyze PolSAR image, such as color features, texture features, spatial relations, etc.
Based on the above basic features, some multiple features of PolSAR data have been constructed to improve the classification performance \cite{Liu2016POL, chen2018polsar}.

For PolSAR image classification tasks, it is also important to design an appropriate classifier, besides the feature extraction.
The supervised classification is a common strategy, which uses enough labeled samples to train the classifiers and determines the class of other samples. Lots of methods have been introduced, including support vector machines, sparse representation, deep learning, etc \cite{chen2017multilayer, xu2018dmil}.

In this paper, we focus on both feature extraction and classifier and propose a PolSAR image classification framework based on polarimetric scattering coding \cite{xu2018pcn} and sparse support matrix machine \cite{zheng2018sparse}. Polarimetric scattering coding is a new feature coding way for PolSAR image and gets a real value matrix. Sparse support matrix machine is a classifier for matrix. 
The combination of these two methods is natural.

\section{METHOD}
\subsection{Polarimetric Scattering Coding}
\label{ssc}

In the PolSAR images, the signals form a $2 \times 2$ complex scattering matrix $S$ to represent the information for one pixel, which relates the incident and the scattered electric fields.
Scattering matrix $S$ can be expressed as

\begin{equation}
\label{eqs1}
S=
\begin{bmatrix}
S_{HH} & S_{HV} \\
S_{VH} & S_{VV}
\end{bmatrix}
\end{equation}

where $S_{HH},S_{HV},S_{VH}$ and $S_{VV}$ are the complex scattering coefficients, $S_{HV}$ is the scattering coefficient of the horizontal(H)
transmitting and vertical(V) receiving polarization.

In the polarimetric scattering coding (PSC) \cite{xu2018pcn}, it assumes that $z=(x+yi)$ is a complex value,
$x$ and $y$ are the real and imaginary parts of $z$ respectively. Polarimetric scattering coding $\varphi$ is shown as follows:



\begin{equation}
\varphi \left ( x+yi \right ) =
\begin{bmatrix}
x & 0 \\
0 & \left | y \right |
\end{bmatrix},
if \; x\ge0 \; and \; y<0\\
\end{equation}

$\varphi$ represents the function of polarimetric scattering coding, when $x>0$, $y<0$.

Because $S$ is a complex matrix, its elements can be written as $S_{HH} = a+bi$, $S_{HV} = c+di$, $S_{VH} = e+fi$, $S_{VV} = g+hi$.
When $a, b, e, h > 0$, $c, d, f, g < 0$. This assumption can take into account the characteristics of the PolSAR data.
\begin{align}
\nonumber \varphi \left (S \right ) &= \varphi \Bigg(\begin{bmatrix}
S_{HH} & S_{HV} \\
S_{VH} & S_{VV}
\end{bmatrix}\Bigg) \\
\nonumber &= \varphi \Bigg(
\begin{bmatrix}
a+bi & c+di \\
e+fi & g+hi
\end{bmatrix}\Bigg) \\
&= \begin{pmat}[{.|}]
a & b                  & 0                  & 0                 \cr
0 & 0                  & \left | c \right | & \left | d \right | \cr \-
e & 0                  & 0                  & h \cr
0 & \left | f \right | & \left | g \right | & 0 \cr
  \end{pmat}
\end{align}

\subsection{Sparse Support Matrix Machine}
The hinge loss enjoys the large margin principle. Support matrix machine (SMM) employs the hinge loss function to get a good classifier, which
takes into account two desirable properties, sparseness and robustness \cite{luo2015support}.
The formulation of the support matrix machine is given as follows:
\begin{equation}
\label{smm}
\begin{split}
\text{arg min}_{W,b}  & \quad \frac{1}{2}tr(W^{T}W)+\tau \left \| W \right \|_{*} \\
& \quad +C\sum_{n}^{i=1}\left \{ 1-y_{i}\left [ tr(W^{T}X_{i})+b \right ] \right \}_{+}
\end{split}
\end{equation}
Where $W$ is the matrix $ \mathbb{R}^{m \times d}$, SMM is based on a penalty function, which is a combination of the squared Frobenius norm $\left \| W \right \|_{2}^{F}$ and the nuclear norm $\left \| W \right \|_{*}$£¬
 $tr(W^{T}W)= vec(W^{T})^{T}vec(W^{T})$ and $tr(W^{T}X_{i})= vec(W^{T})^{T}vec(X_{i})$,
 Thus, the SMM is able to capture the correlation within the input data matrix.

In \cite{zheng2018sparse}, sparse support matrix machine (SSMM) is proposed, which is favored for taking both the intrinsic structure of each input matrix and feature selection into consideration simultaneously.
Both low-rank and sparse constraints on the regression matrix $W$ is imposed . In particular, the objective function of SSMM method is shown as follows:
\begin{equation}
\label{ssmm}
\begin{split}
\text{arg min}_{W,b}  & \quad \gamma \left \| W_{1} \right \|+\tau \left \| W \right \|_{*} \\
& \quad +C\sum_{n}^{i=1}\left \{ 1-y_{i}\left [ tr(W^{T}X_{i})+b \right ] \right \}_{+}
\end{split}
\end{equation}
Where the regularization term on $W$ is a linear combination of $L1$ norm $\left \| W \right \|_{1}$ to control the sparseness.
This method incorporates the hinge loss and constraints on regression matrix $W$ for matrix classification.

Above all, the combination of polarimetric scattering coding and sparse support matrix machine is naturally suitable for PolSAR image classification.

\section{EXPERIMENTS AND ANALYSIS}
In order to compare the performance of vector-based classifiers and matrix-based classifiers, we set two vector-based classifiers, i.e., SVM and sparse SVM (SSVM), as the baseline methods.
We further compare with matrix classifier, such as support matrix machine (SMM).

\subsection{Experiment Data}
In this section, the popular PolSAR image is used to verify the performance of the proposed algorithm. The details are listed in Table \ref{tabf4}. The parameter settings of the proposed method are also discussed. Finally, the results and analysis are given.
\subsection{Data set description}
For our experiments and evaluations, we select a PolSAR
image from an airborne system (NASA/JPL-Caltech AIRSAR).
The information about the PolSAR image is shown below.
The PolSAR image of Flevoland is shown in Fig. \ref{fs4}(a), there are 15 categories in the ground truth map in Fig. \ref{fs4}(b), and the color code is shown in Fig. \ref{fs4}(c).
The spatial resolution is 10 m for 20 MHz. The size of this PolSAR image is 750 $\times$ 1024.
There are 15 kinds of objects to be identified, including stem beans, rapeseed, bare soil, potatoes, beet, wheat2, peas, wheat3, lucerne, barley, wheat, grasses, forest, water and building. These objects are simply written as c1-c15. The numbers of the train and test samples are shown in Table \ref{tabf4}.
\begin{figure}[!htbp]
\begin{minipage}[b]{.54\linewidth}
  \centering
 \centerline{\epsfig{figure= 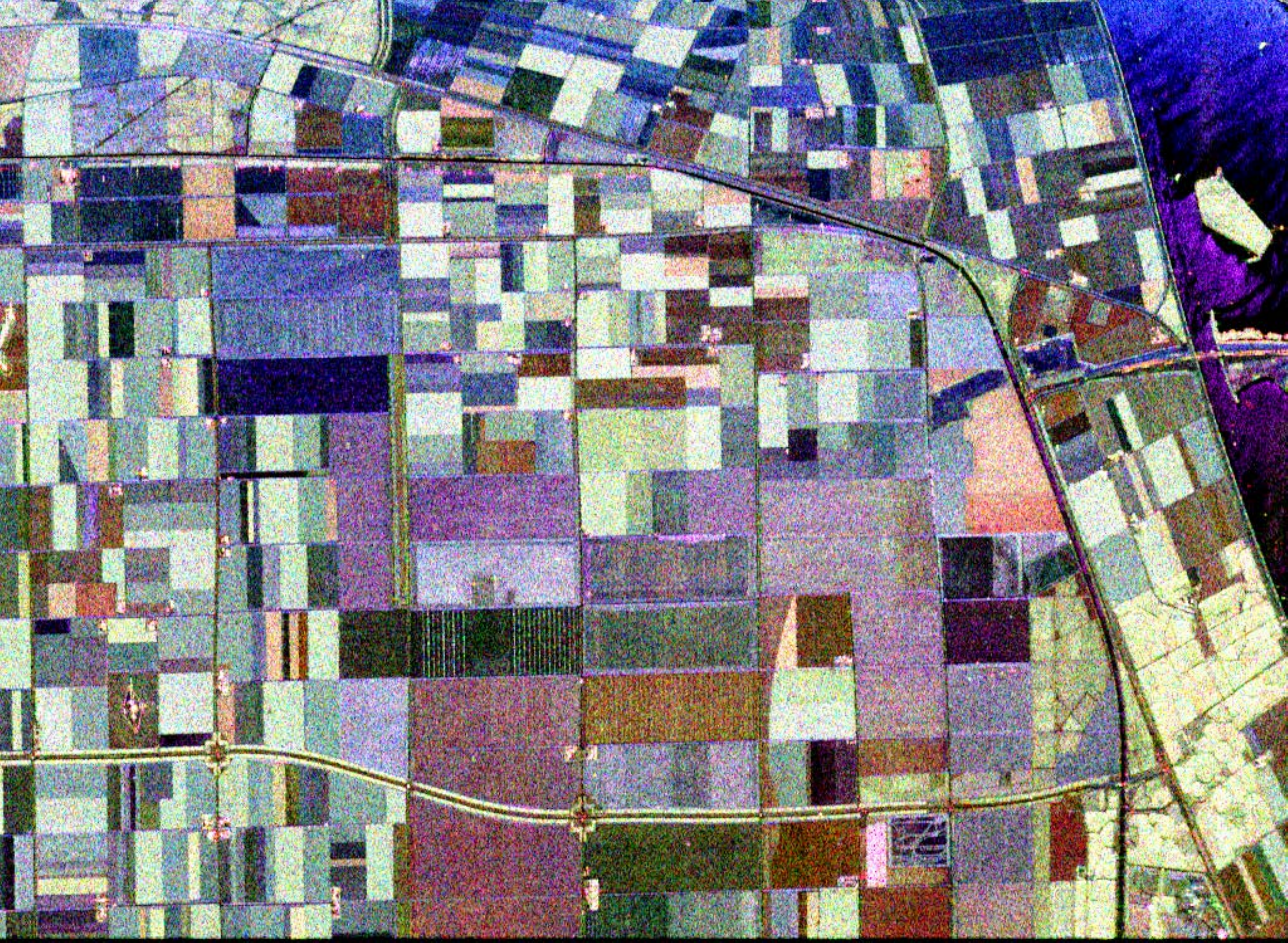,width=4.5cm}}
  \vspace{0.05cm}
  \centerline{(a)}\medskip
\end{minipage}
\begin{minipage}[b]{.40\linewidth}
  \centering
 \centerline{\epsfig{figure= 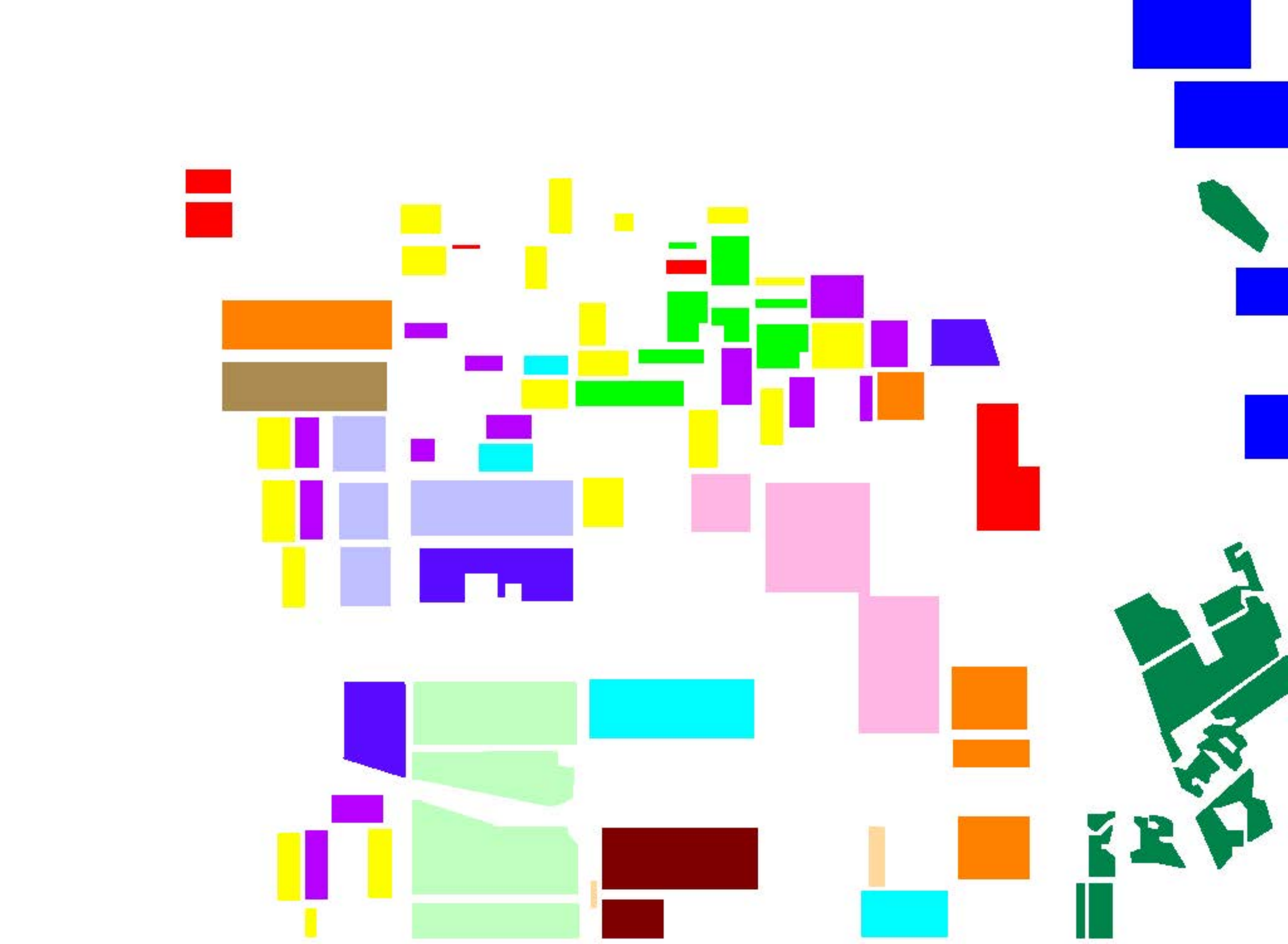,width=4.5cm}}
  \vspace{0.05cm}
  \centerline{(b)}\medskip
\end{minipage}

\begin{minipage}[b]{1\linewidth}
  \centering
 \centerline{\epsfig{figure= 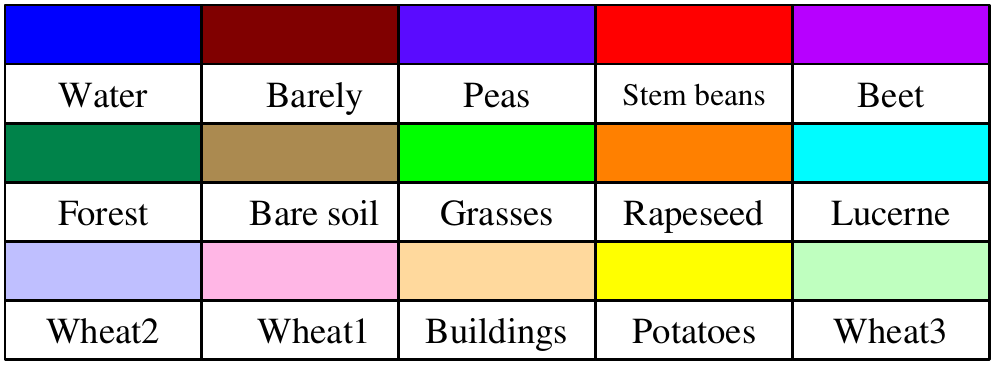,width=8.0cm}}
  \vspace{0.05cm}
  \centerline{(c)}\medskip
\end{minipage}
\caption{Flevoland image and ground truth, AIRSAR. (a) Flevoland image. (b) Ground truth image. (c) The color code.}
\label{fs4}
\end{figure}

\begin{table}[htbp]
\newcommand{\tabincell}[2]{\begin{tabular}{@{}#1@{}}#2\end{tabular}}
\centering
\caption{ \protect\ Land classes and pixels numbers in the PolSAR image. Random selection of 500 samples for each category.}
 \vspace{0.35cm}
\begin{tabular}{c|c|c|c}
\hline
\hline
class code & name &\tabincell{c}{No. of training \\ samples} & \tabincell{c}{No. of testing \\ samples}\\
\hline
1 & Water & 500 & 12732\\
2 & Barely & 500 & 7095\\
3 & Peas & 500 & 9082\\
4 & Stem beans & 500 & 5838\\
5 & Beet & 500 & 9533\\
6 & Forest & 500 & 17544\\
7 & Bare soil & 500 & 4609\\
8 & Grasses & 500 & 6558\\
9 & Rapeseed & 500 & 13363\\
10 & Lucerne & 500 & 9681\\
11 & Wheat2 & 500 & 10659\\
12 & Wheat1 & 500 & 15886\\
13 & Buildings & 500 & 535\\
14 & Potatoes & 500 & 15656\\
15 & Wheat3 & 500 & 21741\\
\hline
\hline
\end{tabular}
  \label{tabf4}
\end{table}

\subsection{Experiment Setting}
In the experiment, there are two parameters $\gamma$ and $C$ to control the trade-off between the regularization terms and the hinge loss.
The two hyperparameters are set to 0.3 and 0.7, respectively.
For the sake of fair comparison, the free parameters of all competitive methods are carefully tuned in order to obtain their best classification results.
For vector-based classifiers, i.e., SVM and SSVM,
traditional polarimetric features are extracted for comparing. A common 22-dimensional feature vector is used as the compared polarimetric feature, which includes the upper right element's absolute value of the 3$\times$3 polarimetric coherency matrix , the upper right element's absolute value of the 3$\times$3 polarimetric covariance matrix, three components of Pauli decomposition, three components of Freeman decomposition, and four components of Yamaguchi decomposition, expressed as PF22.

\subsection{Results and Analysis}
\begin{figure}[!htb]
  \centering
\begin{minipage}[b]{0.48\linewidth}
 \centerline{\epsfig{figure= 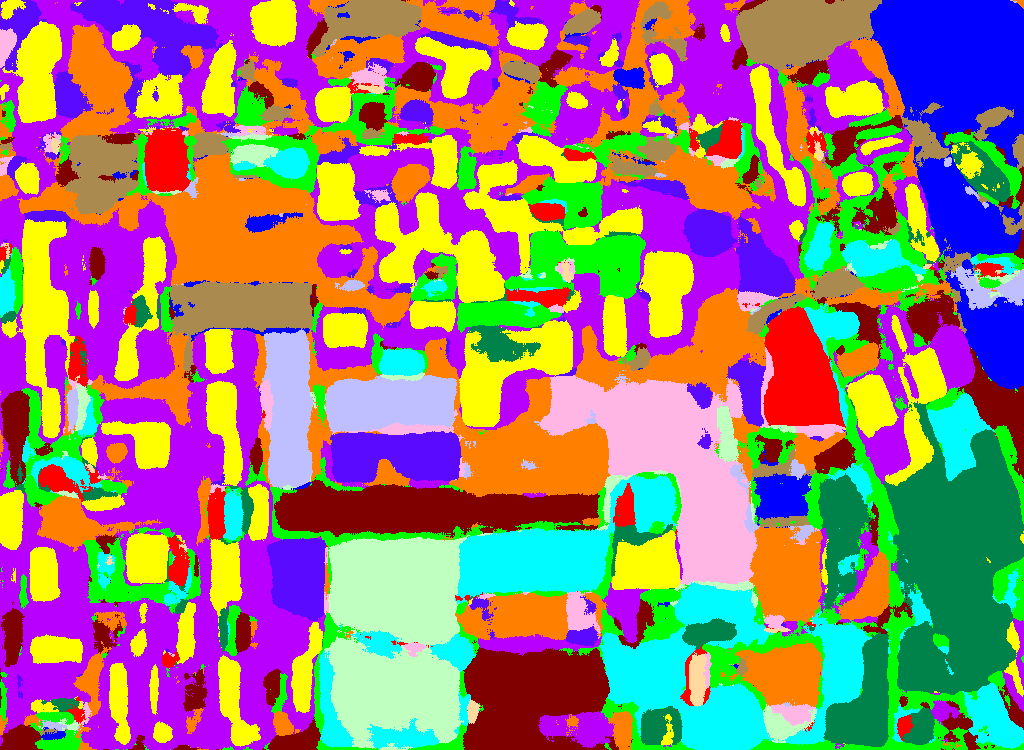,width=4.1cm}}
  \centerline{(a) }\medskip
\end{minipage}
\begin{minipage}[b]{0.51\linewidth}
 \centerline{\epsfig{figure= 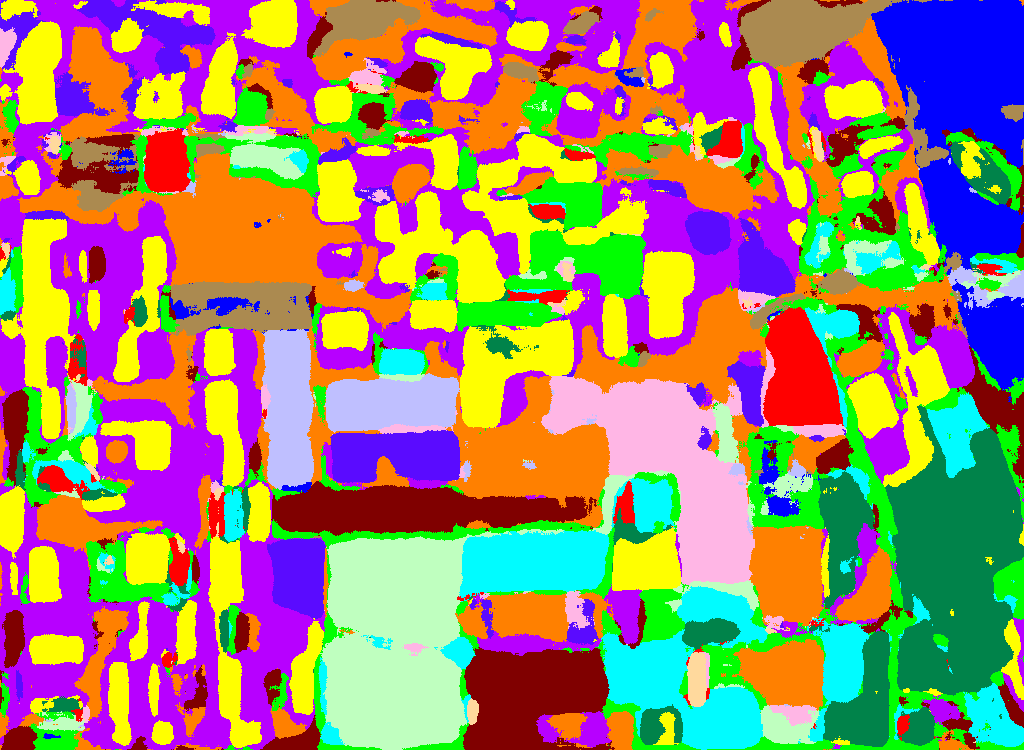,width=4.1cm}}
  \centerline{(b) }\medskip
\end{minipage}

\begin{minipage}[b]{0.48\linewidth}
 \centerline{\epsfig{figure= 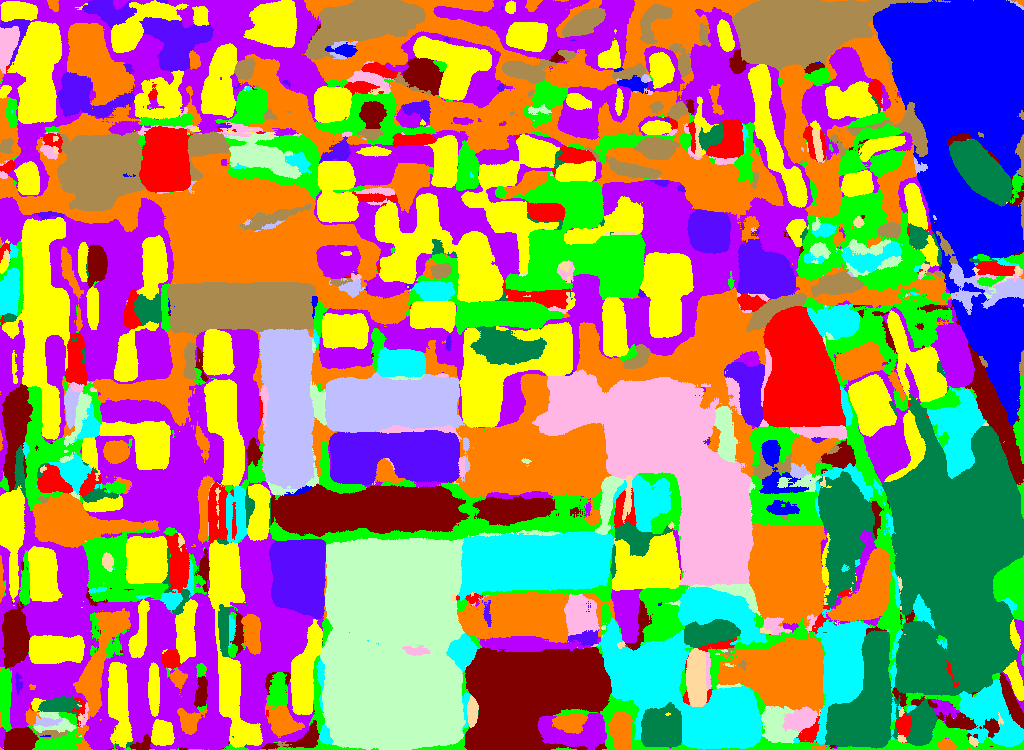,width=4.1cm}}
  \centerline{(c) }\medskip
\end{minipage}
\begin{minipage}[b]{0.51\linewidth}
 \centerline{\epsfig{figure= 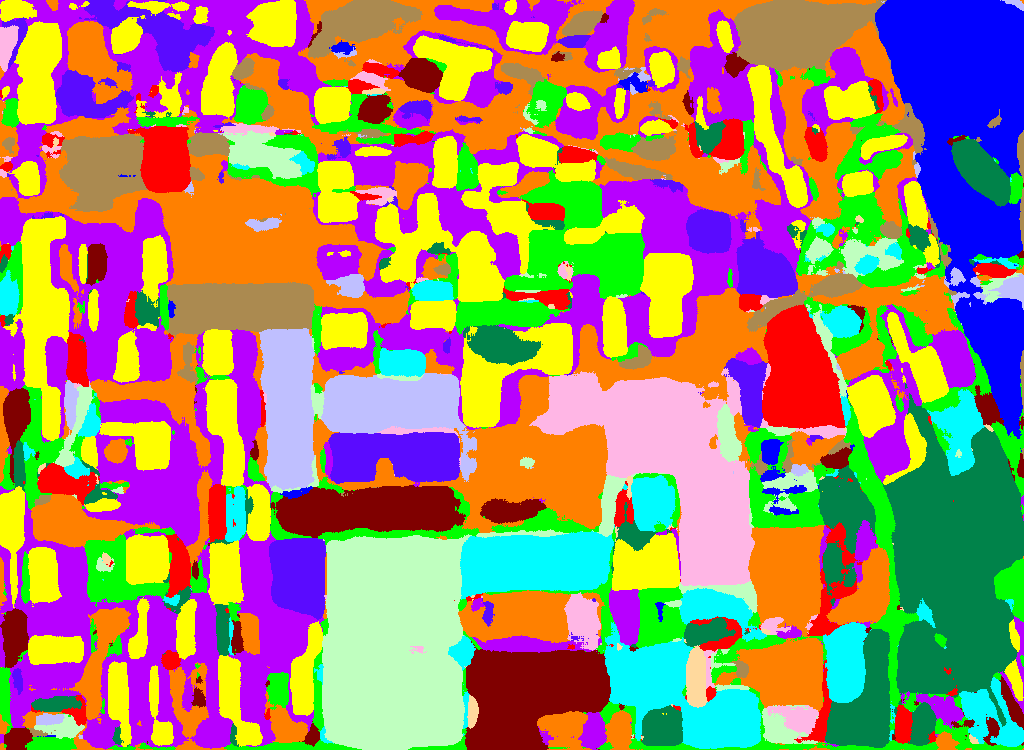,width=4.1cm}}
  \centerline{(d) }\medskip
\end{minipage}
\caption{The classification maps on the PolSAR image. (a)-(d) : PF22-SVM, PF22-SSVM, PSC-SMM, PSC-SSMM.}
\label{fs4rs}
\end{figure}

The experimental results are shown in Fig. \ref{fs4rs} and Table \ref{fs4tb}.
Fig. \ref{fs4rs} shows the classification maps of the PolSAR image.
The original image is shown in Fig. \ref{fs4}, which contains 15 kinds of objects.
It is difficult to recognize the samples when facing the complex characteristics of interclass and intraclass.
For instance, wheat1 wheat2 and wheat3 are similar and indistinguishable, the wheat3 is often regard as wheat1 and wheat2 by classifier. As is shown in Fig. \ref{fs4rs}(a)-(b), Fig. \ref{fs4rs}(a) is remarkable, Fig. \ref{fs4rs}(b) is relatively few.
In the proposed method, the classification effect has been greatly improved, which is shown in Fig. \ref{fs4rs}(d).
Similarly, some potatoes are wrongly classified as peas, but the proposed method can give a correct judgement.

In Table \ref{fs4tb}, the above experimental phenomena can be seen accurately through the value of overall
accuracy (OA), average accuracy (AA) and kappa coefficient
(Kappa). The classification accuracy can be improved obviously.

We can see that the proposed approach outperforms the compared methods. It indicates that the encoded data through polarimetric scattering coding is easier to be identified and distinguished. At the same time, we can find that the sparse support matrix machine has a better classification performance than support matrix machine.

\begin{table}[!htp]
\centering

\caption{Classification accuracy (\%) of the PolSAR image.}
 \vspace{0.35cm}
\setlength{\tabcolsep}{0.8mm}{
\begin{tabular}{c|c|c|c|c}
\hline
\hline
Method & PF22-SVM & PF22-SSVM & PSC-SMM & PSC-SSMM \\
\hline
c1 & 81.33 & 86.35 & 88.95 & 98.06 \\
c2 & 82.79 & 86.09 & 87.13 & 94.08 \\
c3 & 87.31 & 85.13 & 88.21 & 92.34 \\
c4 & 88.20 & 88.77 & 89.43 & 95.76 \\
c5 & 82.23 & 88.55 & 93.24 & 96.33 \\
c6 & 81.42 & 93.88 & 95.88 & 92.12 \\
c7 & 81.18 & 93.64 & 89.46 & 97.66 \\
c8 & 86.38 & 90.04 & 95.85 & 93.36 \\
c9 & 85.21 & 88.99 & 92.76 & 90.81 \\
c10 & 85.68 & 91.04 & 92.43 & 90.87 \\
c11 & 88.52 & 86.06 & 92.22 & 92.12 \\
c12 & 80.35 & 93.21 & 87.23 & 93.56 \\
c13 & 87.44 & 93.33 & 92.87 & 93.54 \\
c14 & 86.34 & 87.11 & 93.10 & 91.23 \\
c15 & 81.72 & 88.31 & 95.65 & 91.56 \\
\hline
AA & 84.46 & 89.36 & 91.62 & 93.56 \\
OA & 84.83 & 91.51 & 91.99 & 95.65 \\
Kappa & 0.826 & 0.875 & 0.908 & 0.935 \\
\hline
\hline
\end{tabular}}
\label{fs4tb}
\end{table}

\section{CONCLUSIONS}
In this paper, a novel classification framework is proposed for PolSAR image, which is based on polarimetric scattering coding and sparse support matrix machine. The polarimetric scattering coding can transfer the complex-value scattering matrix to a real value matrix. Then, we introduce the sparse support matrix machine, the real value matrix can be fed into the model directly. In the experiment, the proposed method gives a better result.
The combination of polarimetric scattering coding and sparse support matrix machine is a novel and effective way for PolSAR image classification.
\section{Acknowledgment}
The authors would like to thank the anonymous reviewers for their helpful comments.
The authors would also like to thank the NASA/JPL-Caltech and Canadian Space Agency for providing the polarimetric AIRSAR data.

\bibliographystyle{IEEEbib}
\bibliography{refs}

\end{document}